# Humanoid Momentum Estimation Using Sensed Contact Wrenches

Nicholas Rotella[1], Alexander Herzog[2], Stefan Schaal[1,2] and Ludovic Righetti[2]

*Abstract*—This work presents approaches for the estimation of quantities important for the control of the momentum of a humanoid robot. In contrast to previous approaches which use simplified models such as the Linear Inverted Pendulum Model, we present estimators based on the momentum dynamics of the robot. By using this simple yet dynamically-consistent model, we avoid the issues of using simplified models for estimation. We develop an estimator for the center of mass and full momentum which can be reformulated to estimate center of mass offsets as well as external wrenches applied to the robot. The observability of these estimators is investigated and their performance is evaluated in comparison to previous approaches.

## I. INTRODUCTION

Recent work has shown the utility of controlling momentum in order to stabilize humanoid robots and generate dynamic motion [1],[2],[3],[4]. These approaches, however, are limited on real systems due to the lack of accurate estimates of momentum. Even traditional approaches using simplified models such as the Linear Inverted Pendulum Model (LIPM) for planning rely on accurate center of mass (COM) estimates to achieve good tracking. Few papers have been published on humanoid state estimation approaches which address these issues. This work introduces several different estimators which address the shortcomings of previous approaches by using dynamics consistent with the full robot model to estimate the COM, linear and angular momentum as well as COM and momentum offsets and external wrenches.

Kwon et al. [5] recognized the challenge of performing ZMP preview control using inaccurate COM information. Their solution was a Kalman Filter (KF) which fused a LIPM-based process model and the expected motion of the COM, computed from the previewed ZMP and resolved into joint angles. This relied heavily on the expected ZMP, making it suitable only for open-loop walking on simple terrain. Endeffector force/torque (F/T) sensors were used but the simplified model limited their utility. Rotational motion (angular momentum) was ignored.

Stephens [6] investigated simplified models for estimation of COM position/velocity, center of pressure (COP), COM offsets and external forces. He introduced filters based on the LIPM, analyzed their observability and demonstrated their performance. However, the use of a LIPM-based process model compromises the filter optimality which can lead to delays and even destabilize control. In addition, the observability of a simplified model is not the same as that of the original system. Third, use of the COP limits these estimators to planar terrain. Finally, this work does not consider estimation of angular momentum.

Xinjilefu et al. [7] investigated whether simplified models are sufficient for estimation. They fused LIPM dynamics with COM and COP measurements in one estimator and a planar, five-link model with joint angle measurements and inertial measurements from an IMU in another. The estimators were evaluated on simulated data with added biases. However, biases were not modeled explicitly; instead, noise parameters were tuned to account for them. No observability analysis was performed, making the filters difficult to analyze; they are difficult to compare since they serve different purposes.

Hashlamon et al. [8] presented a COM and COP estimator which does not require F/T sensors. Contact wrenches and COPs were determined from IMU acceleration by solving an optimization problem using contact constraints. However, they assumed the IMU measures COM acceleration and that the angular momentum is constant. They then presented a LIPM-based KF which uses the computed COP as an input to the LIPM dynamics. The resulting filter, despite providing COP estimates without using F/T sensors, demonstrates poor performance likely due to its strong assumptions.

Xinjilefu et al. [9] developed an optimization-based estimator which uses the full dynamics and measurements from many sensors. Both states and controls are estimated in a quadratic program (QP) and sensors can be easily integrated since the dynamics are linear in terms of the state. Additionally, constraints can be enforced whereas this is not straightforward to do in a KF. However, this approach relies heavily on the full robot dynamics (namely link inertias) which we usually do not have good estimates of. Further, offsets in the COM and momentum are not considered.

Most previous studies used simplified models to avoid using the full dynamics. However, for linear systems it can be shown [10] that an inaccurate process model can lead to unstable estimation error dynamics; stability proofs assume exact plant cancellation. Further, an inaccurate model can destabilize both the controller and observer. This is because the separation principle - which says that the controller and observer can be designed independently - no longer holds. Of course, since the true dynamics are nonlinear, these theorems can only be applied qualitatively to inform design. This suggests avoiding simplified models when possible; we thus present estimators which use the momentum dynamics.

This research was supported in part by National Science Foundation grants IIS-1205249, IIS-1017134, CNS-0960061, EECS-0926052, the DARPA program on Autonomous Robotic Manipulation, the Office of Naval Research, the Okawa Foundation, the Max-Planck-Society and the European Research Council under the European Union's Horizon 2020 research and innovation programme (grant agreement No 637935). Any opinions, findings, and conclusions or recommendations expressed in this material are those of the author(s) and do not necessarily reflect the views of the funding organizations.

[1]Computational Learning and Motor Control Lab, University of Southern California, Los Angeles, California.
[2]Autonomous Motion Department, Max Planck Institute for Intelligent Systems, Tuebingen, Germany.

Also known as the Newton-Euler equations, they appear in the full rigid body dynamics as the equations describing the unactuated portion of the state; they are *reduced*, not simplified, dynamics. Motivated by issues observed on our SARCOS humanoid, we develop estimators which:

- Filter noisy kinematics-based COM and momentum measurements without inducing significant delays (Momentum Estimator, Section III-A).
- Estimate configuration-dependent COM and momentum offsets caused by inaccurate link model parameters (Offset Estimator and COP-Based Offset Estimator, Sections III-B and III-C).
- Estimate a time-varying external wrench applied to the COM (External Wrench Estimator, Section III-D).

We investigate the observability of each of the presented estimators and compare their performance against corresponding LIPM-based estimators proposed in [6].

## II. MODELS OF ROBOT DYNAMICS

In recent work, we have implemented a momentum controller derived as a state feedback controller using LQR design and realized using a hierarchical QP-based inverse dynamics solver [4]. We thus design a momentum estimator to use with this controller based on the same dynamics, given below.

$$\dot{c} = \frac{1}{m} l \qquad (1)$$

$$\dot{l} = \sum_{i=1}^{M} (F_i + w_{F_i}) + mg \qquad (2)$$

$$\dot{k} = \sum_{i=1}^{M} (p_i - c) \times (F_i + w_{F_i}) + \sum_{i=1}^{M} (\tau_i + w_{\tau_i}) \qquad (3)$$

Here $c$ is the COM, $l$ and $k$ are the linear and angular momentum respectively, $p_i$ is the $i^{th}$ point of contact ($M$ in total), $F_i$ and $\tau_i$ are the force and torque at the $i^{th}$ contact respectively, $m$ is the total mass of the robot and $g$ is the gravity vector. The vectors $w_{F_i}$ and $w_{\tau_i}$ represent additive white Gaussian force and torque noise processes, respectively (with standard deviations $q_F$ and $q_\tau$).

These dynamics are preferable to the full dynamics for prediction because they avoid using inaccurate link inertia models and instead use only the total mass, which can be measured. Additionally, they can be linearized easily. Qualitatively, an estimator which uses the true momentum dynamics should interact with a momentum controller more favorably than one which uses simplified dynamics.

The estimators in [6] are derived from the LIPM dynamics

$$\ddot{x} = \frac{g}{z_c}(x - x_{COP}) \qquad (4)$$

where $x$ denotes the COM position, $z_c$ is the (constant) height of the COM and $x_{COP}$ denotes the COP position. The dynamics of the $y$ direction are exactly the same since the LIPM is decoupled. This equation can be derived from (2) and (3) by replacing $p_i$ with the COP, neglecting rotation and constraining motion to a plane. For use with less-dynamic controllers, an estimator based on (4) may prove sufficient.

## III. ESTIMATORS

In this section, we introduce four estimators based on the momentum dynamics, each serving a different purpose. We choose to implement these using Extended Kalman Filters (EKFs), however the observability results of Section IV hold for any implementation. Each estimator's nonlinear process dynamics are used for prediction but are linearized for the update step into the state-space dynamics $\dot{x} = Ax + Lw$ and the measurement $y = Cx + v$ (with $x$ being the state and $w$ and $v$ the process and measurement noise vectors). We specify estimators by their nonlinear process and measurement models as well as the continuous-time Jacobians $A$, $L$ and $C$ resulting from linearization.

### A. Momentum Estimator

On our humanoid, joint velocities are computed by differentiating noisy potentiometer signals; momentum is computed from kinematics using these noisy velocities and base information from a base-state estimator developed in previous work [11]. We currently filter momentum using second-order Butterworth low-pass filters but this introduces delays. By integrating low-noise measured contact wrenches and using the noisy kinematics computations as measurements, we can filter momentum without inducing delays.

We choose the state $x = [c, l, k]$ so the process model is (1)-(3) and the measurement model is $y = [c, k]$; we measure the COM and angular momentum computed from base information, kinematics and inertias. These measurements have noise standard deviations $r_c$ and $r_k$, respectively. Although inaccurate inertias are used to compute $k$, we demonstrate robustness to model errors in Section V. The dynamics are

$$A = \begin{bmatrix} 0 & \frac{1}{m}I & 0 \\ 0 & 0 & 0 \\ \sum \bar{F}_i^\times & 0 & 0 \end{bmatrix}, \quad L_i = \begin{bmatrix} 0 & 0 \\ I & 0 \\ (\bar{p}_i - \bar{c})^\times & I \end{bmatrix}$$

$$C = \begin{bmatrix} I & 0 & 0 \\ 0 & 0 & I \end{bmatrix}$$

where $a^\times$ denotes the cross product matrix formed from vector $a$. The noise Jacobian $L$ is formed by concatenating $L_i$ horizontally for all $i = 1 \ldots M$ contacts. $\bar{F}_i$, $\bar{p}_i$ and $\bar{c}$ are the measured force and the foot and COM positions, respectively, treated as constants over a single timestep.

### B. Offset Estimator

The kinematic model of a robot is often imprecise. Here, we assume the total mass $m$ is known but that incorrect link masses and COM locations contribute to a configuration-dependent COM offset. These errors also create offsets in momentum; we thus extend the state to $x = [c, l, k, \Delta c, \Delta l]$ where $\Delta c$ and $\Delta l$ are the COM and linear momentum offset vectors, respectively. In theory, there is also an offset $\Delta k$ but this is unobservable (see Section IV) so we choose not to estimate it. Since the offset dynamics are unknown, we assume random walks so the prediction dynamics are (1)-(3) plus $\Delta \dot{c} = w_{\Delta c}$ and $\Delta \dot{l} = w_{\Delta l}$ where $w_{\Delta c}$ and $w_{\Delta l}$ are additive Gaussian white noise processes with standard

deviations $q_{\Delta c}$ and $q_{\Delta l}$. The measurement model is

$$y = \begin{bmatrix} c + \begin{bmatrix} \Delta c_x \\ \Delta c_y \\ 0 \end{bmatrix} \\ l + \Delta l \\ k \end{bmatrix}$$

where the linear momentum measurement noise has standard deviation $r_l$. This equation indicates that the kinematics-based COM and linear momentum measurements contain offsets. It is shown in Section IV that $\Delta c_z$ is unobservable and is thus ignored.[1] The linearized dynamics are given by

$$A = \begin{bmatrix} 0 & \frac{1}{m}I & 0 & 0 & 0 \\ 0 & 0 & 0 & 0 & 0 \\ \sum \bar{F}_i^\times & 0 & 0 & 0 & 0 \\ 0 & 0 & 0 & 0 & 0 \\ 0 & 0 & 0 & 0 & 0 \end{bmatrix}, L_i = \begin{bmatrix} 0 & 0 \\ I & 0 \\ (\bar{p}_i - \bar{c})^\times & I \\ 0 & 0 \\ 0 & 0 \end{bmatrix}$$

$$C = \begin{bmatrix} I & 0 & 0 & I_2 & 0 \\ 0 & I & 0 & 0 & I \\ 0 & 0 & I & 0 & 0 \end{bmatrix}$$

where $I_2 = \text{diag}(1,1,0)$ since we ignore $\Delta c_z$. Because the dynamics of $\Delta c_z$ affect the estimate of $l_z$, it is best to ignore this offset and accept that it cannot be observed. This is why we write the measurement this way. Note that we do not explicitly consider link inertia errors, however these would contribute only to the unobservable offset $\Delta k$.

### C. COP-Based Offset Estimator

In the above estimator, $\Delta c_x$ and $\Delta c_y$ are observable due to the dynamics of the angular momentum measurement (see Section IV for details). However, this measurement is subject to an offset $\Delta k$. While ignored above, this leads to degraded performance for significant modeling errors. We add the COP measurement to give us force-based information about the COM which is accurate despite modeling errors. This comes at the disadvantage of assuming coplanar contacts.

We use the same state and add the COP measurement (with noise standard deviation $r_{COP}$). We keep the measurement of $k$ and increase $r_k$ relative to $r_{COP}$ in order to filter angular momentum while relying primarily on the COP to render $\Delta c_x$ and $\Delta c_y$ observable. The measurement model becomes nonlinear with the addition of the measurement

$$y_{COP,x} = c_x - \frac{1}{\sum F_{i,z}} \left( c_z \sum F_{i,x} + \dot{k}_y \right)$$
$$y_{COP,y} = c_y - \frac{1}{\sum F_{i,z}} \left( c_z \sum F_{i,y} - \dot{k}_x \right)$$

where $\dot{k} = \sum(p_i - c) \times F_i + \sum \tau_i$. The measurement Jacobian

[1] Unobservable states can drift arbitrarily; when coupled to other states through their dynamics, this drift affects their estimation. Because of this, adding an unobservable offset $\Delta k$ would corrupt the information about $\Delta c$ which is provided by the angular momentum measurement. We are better off incorrectly assuming that $\Delta k = 0$ and accepting that our estimates of $\Delta c$ may be inaccurate depending on the severity of the modeling errors. The filter of Section III-C addresses this.

is

$$C = \begin{bmatrix} I & 0 & 0 & I_2 & 0 \\ 0 & I & 0 & 0 & I \\ 0 & 0 & I & 0 & 0 \\ H & 0 & 0 & 0 & 0 \end{bmatrix}, \quad H = \begin{bmatrix} 2 & 0 & \frac{-2\sum \bar{F}_{i_x}}{\sum F_{i_z}} \\ 0 & 2 & \frac{-2\sum \bar{F}_{i_x}}{\sum F_{i_z}} \end{bmatrix}$$

where $H$ is the Jacobian relating the COM to the COP.

### D. External Wrench Estimator

It is often the case that F/T sensors drift or disturbances are applied to the robot. We begin with the Momentum Estimator state and introduce an external wrench acting at the COM which encapsulates these errors. The state becomes $x = [c, l, k, \Delta F, \Delta \tau]$ where $\Delta F, \Delta \tau$ are the external force and COM torque (torque computed about the COM). Again, we assume random walks for each so the dynamics are

$$\dot{c} = \frac{1}{m}l$$
$$\dot{l} = \sum_{i=1}^M (F_i + w_{F_i}) + mg + \Delta F$$
$$\dot{k} = \sum_{i=1}^M (p_i - c) \times (F_i + w_{F_i}) + \sum_{i=1}^M (\tau_i + w_{\tau_i}) + \Delta \tau$$
$$\Delta \dot{F} = w_{\Delta F}$$
$$\Delta \dot{\tau} = w_{\Delta \tau}$$

where $w_{\Delta F}$ and $w_{\Delta \tau}$ have standard deviations $q_{\Delta F}$ and $q_{\Delta \tau}$. The measurement model is $y = [c, k]$ as in the original Momentum Estimator. The linearized dynamics are

$$A = \begin{bmatrix} 0 & \frac{1}{m}I & 0 & 0 & 0 \\ 0 & 0 & 0 & I & 0 \\ \sum F_i^\times & 0 & 0 & 0 & I \\ 0 & 0 & 0 & 0 & 0 \\ 0 & 0 & 0 & 0 & 0 \end{bmatrix}, L_i = \begin{bmatrix} 0 & 0 \\ I & 0 \\ (\bar{p}_i - \bar{c})^\times & I \\ 0 & 0 \\ 0 & 0 \end{bmatrix}$$

$$C = \begin{bmatrix} I & 0 & 0 & 0 & 0 \\ 0 & 0 & I & 0 & 0 \end{bmatrix}$$

## IV. OBSERVABILITY ANALYSIS

The presented estimators have nonlinear process models and, where the COP is used, nonlinear measurement models. We investigate the observability of each estimator by forming the nonlinear observability matrix [12], denoted $O$. As in the linear case, the state is observable if $O$ has full rank; unobservable state combinations are parameterized by the nullspace of $O$. Given a system with nonlinear process model $\dot{x} = f(x)$ and measurement model $y = h(x)$,

$$O = \begin{bmatrix} \nabla h(x) \\ \nabla(\nabla h \bullet f(x)) \\ \nabla(\nabla(\nabla h \bullet f(x)) \bullet f(x)) \\ \vdots \end{bmatrix}$$

where $\nabla$ denotes the Jacobian with respect to $x$. In the linear case, $\nabla h = C$ and $f(x) = Ax$ so $\nabla(\nabla h \bullet f(x)) = \nabla(CAx) = CA$, $\nabla(\nabla(\nabla h \bullet f(x)) \bullet f(x)) = CA^2$ and so on. Unlike in the linear case, there is no condition limiting the size of $O$; however, only a finite number of derivatives usually need be taken before successive rows become zero (and thus no longer affect the rank). Essentially, $O$ illustrates that states are observable because they are measured directly or because they appear in the dynamics (of some order) of a state which is measured directly.

Note that since $O$ is state-dependent in general, this procedure investigates *local* observability. We cannot list every point at which $O$ becomes rank-deficient but we will highlight relevant cases. Due to space constraints, we present the observability matrix for each estimator without proof; they can be derived by straightforward computation.

### A. Momentum Estimator

The nonlinear observability matrix is

$$O = \begin{bmatrix} I & 0 & 0 \\ 0 & 0 & I \\ 0 & \frac{1}{m}I & 0 \\ \sum F_i^\times & 0 & 0 \\ 0 & \frac{1}{m}\sum F_i^\times & 0 \end{bmatrix}$$

This matrix has full rank so the state is observable. We could add a kinematics-based linear momentum measurement but it would be redundant since it is computed from the same sensors as the COM. Since linear momentum is a function of noisy velocities, it is subject to considerable noise; it is better to leave it out and let differentiation occur in the filter.

### B. Offset Estimator

The observability matrix is given below; we include the full $\Delta c$ in the state to prove that it cannot be observed.

$$O = \begin{bmatrix} I & 0 & 0 & I^* & 0 \\ 0 & I & 0 & 0 & I \\ 0 & 0 & I & 0 & 0 \\ 0 & \frac{1}{m}I & 0 & 0 & 0 \\ \sum F_i^\times & 0 & 0 & 0 & 0 \\ 0 & \frac{1}{m}\sum F_i^\times & 0 & 0 & 0 \end{bmatrix}$$

This matrix is rank deficient; the unobservable subspace is a linear combination of the COM and COM offset. This occurs because the skew-symmetric matrix $\sum F_i^\times$ has at most rank two. In theory, this means we could observe any two directions of the COM offset, effectively replacing $I^*$ in $O$. However, when the robot is stationary on flat ground,

$$\sum F_i^\times = \begin{bmatrix} 0 & -mg & 0 \\ mg & 0 & 0 \\ 0 & 0 & 0 \end{bmatrix}$$

forcing the $z$ direction to be unobservable. Either the $x$ or $y$ direction can be made unobservable in its place only when $\sum F_{i,x}$ or $\sum F_{i,y}$ is nonzero. This occurs only when an force acts on the robot in these directions at the COM (when the robot accelerates or is on a slope so gravity affects $x$ and/or $y$). For this reason, we set $I^* = I_2 = \text{diag}(1,1,0)$ so that $\Delta c_x$ and $\Delta c_y$ are observable. Note that $\sum F_i = 0$ for a robot in flight phase; in this case, $\Delta c$ is completely unobservable.

### C. COP-Based Offset Estimator

The nonlinear observability matrix is

$$O = \begin{bmatrix} I & 0 & 0 & I_2 & 0 \\ 0 & I & 0 & 0 & I \\ 0 & 0 & I & 0 & 0 \\ H & 0 & 0 & 0 & 0 \\ 0 & \frac{1}{m}I & 0 & 0 & 0 \\ \sum F_i^\times & 0 & 0 & 0 & 0 \\ 0 & \frac{1}{m}\sum F_i^\times & 0 & 0 & 0 \\ 0 & \frac{1}{m}H & 0 & 0 & 0 \end{bmatrix}$$

Observability is the same as for the Offset Estimator; however, the COP provides information about the COM derived from F/T sensors rather than from kinematics. This improves offset estimation in the case when the unmodeled angular momentum offset is large. This makes the important point that adding redundant measurements can affect performance even though observability is unchanged.

### D. External Wrench Estimator

The nonlinear observability matrix is

$$O = \begin{bmatrix} I & 0 & 0 & 0 & 0 \\ 0 & 0 & I & 0 & 0 \\ 0 & \frac{1}{m}I & 0 & 0 & 0 \\ \sum F_i^\times & 0 & 0 & 0 & I \\ 0 & 0 & 0 & \frac{1}{m}I & 0 \\ 0 & \frac{1}{m}\sum F_i^\times & 0 & 0 & 0 \\ 0 & 0 & 0 & \frac{1}{m}\sum F_i^\times & 0 \end{bmatrix}$$

This matrix has full rank - since the external force and torque appear in the momentum dynamics, and since momentum is observable through the COM and angular momentum measurements, the external wrench is observable.

### E. Offset and External Wrench Estimator

If we extend the Offset Estimator state so that $x = [c, l, k, \Delta c, \Delta l, \Delta F, \Delta \tau]$ then we obtain the matrix

$$O = \begin{bmatrix} I & 0 & 0 & I^* & 0 & 0 & 0 \\ 0 & I & 0 & 0 & I & 0 & 0 \\ 0 & 0 & I & 0 & 0 & 0 & 0 \\ 0 & \frac{1}{m}I & 0 & 0 & 0 & 0 & 0 \\ 0 & 0 & 0 & 0 & 0 & I & 0 \\ \sum F_i^\times & 0 & 0 & 0 & 0 & 0 & I \\ 0 & 0 & 0 & 0 & 0 & \frac{1}{m}I & 0 \\ 0 & \frac{1}{m}\sum F_i^\times & 0 & 0 & 0 & 0 & 0 \\ 0 & 0 & 0 & 0 & 0 & \frac{1}{m}\sum F_i^\times & 0 \end{bmatrix}$$

This matrix is rank deficient; its nullspace is a linear combination of $c$, $\Delta c$ and $\Delta \tau$. This implies that we can never observe both the offsets and an external wrench simultaneously. However, if the external force is applied at the COM then $\Delta \tau = 0$ and the state (except $\Delta c_z$) becomes observable.

## V. RESULTS

All results presented in this work were obtained in the SL simulation environment [13]. White Gaussian noise was generated based on data from our SARCOS humanoid robot, discretized at the filtering frequency and added to simulated sensor outputs. Noise having standard deviation $q_\theta$ was added to joint angles (and propagated to joint velocities through numerical differentiation) as well as to endeffector F/T sensor signals. Table I below lists the values of the standard deviations of the simulated noise processes.

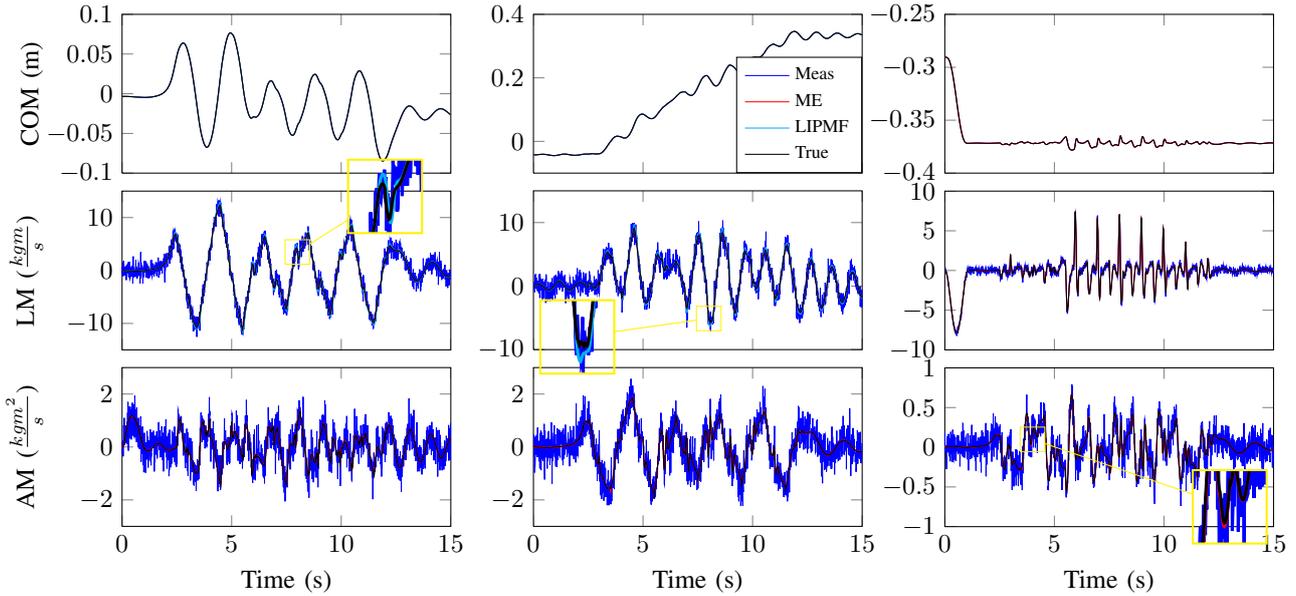

Fig. 1: Estimation of $COM$ (top row), linear momentum (middle row) and angular momentum (bottom row). Inset zoomed-in views (3x magnification) show finer details and demonstrate superior performance of the ME.

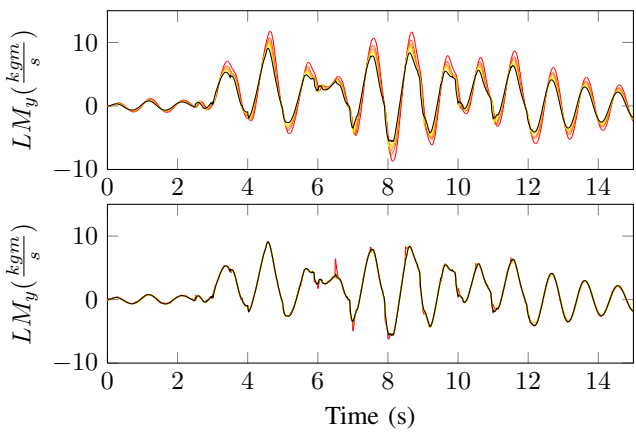

Fig. 2: Estimation of linear momentum for increasing frequencies. Top: LIPM Filter. Bottom: Momentum Estimator.

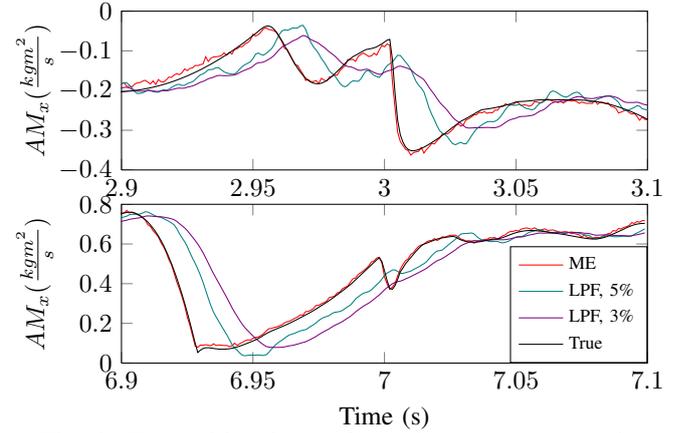

Fig. 3: Zoomed-in views of angular momentum estimates versus low-pass filtered kinematics measurements.

TABLE I: Simulated sensor noise standard deviations. Corresponding values for 1kHz sampling rate are shown.

|  | Continuous | Discrete ($1kHz$) |
| --- | --- | --- |
| $q_\theta$ | $0.00000316 rad/\sqrt{Hz}$ | $0.0001 rad$ |
| $q_F$ | $0.06325 N/\sqrt{Hz}$ | $2N$ |
| $q_\tau$ | $0.00316 Nm/\sqrt{Hz}$ | $0.1 Nm$ |

In this and subsequent sections, we refer to the four filters based on the momentum dynamics as the ME (Momentum Estimator), OE (Offset Estimator), COE (COP-Based Offset Estimator) and EWE (External Wrench Estimator). Process noise parameters were set using the values in Table I. All other noise parameters were tuned for each filter and are summarized in Table II. Note that measurement noise standard deviations are specified in discrete time. All filters based on the LIPM are referred to as LIPMF in this section and denote the corresponding filter and noise parameters introduced in [6].

Estimation was performed during a $15s$ ZMP preview control-based walking task [14] having single and double support phases lasting $0.5s$ each and a forward motion of $5cm$ per step for 10 steps. Since this is a relatively-dynamic gait, the contacts created are often not completely flat and subject to impulsive contact wrenches; this was desired in order to test the estimators with realistic contact switching. Ideal base state estimation was assumed for simplicity but estimators were additionally verified using a base state estimator [11] subject to simulated IMU sensor noise. Estimation and data recording were performed at $1kHz$ unless noted.

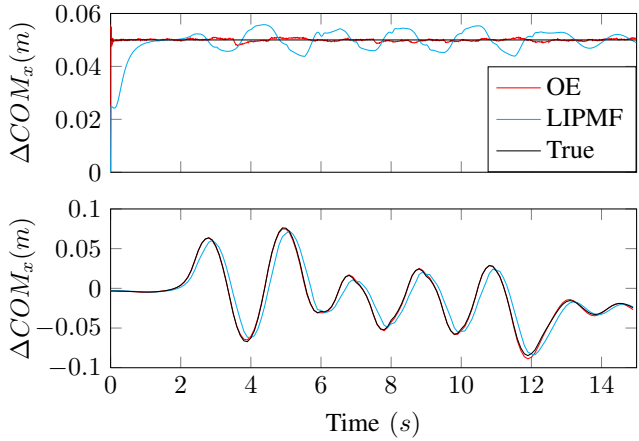

Fig. 4: Estimation of a 5cm COM offset while stationary (top) and estimation of a configuration-dependent COM offset during the $15s$ walking task (bottom).

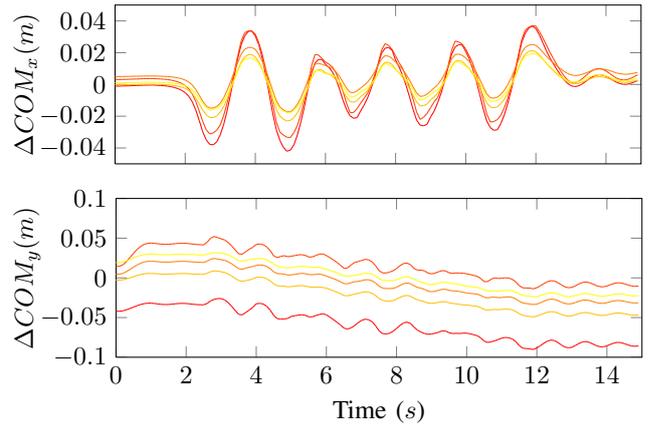

Fig. 5: Time-varying COM offsets for different $n$ (representing different degrees of modeling error) in $x$-direction (top) and $y$-direction (bottom).

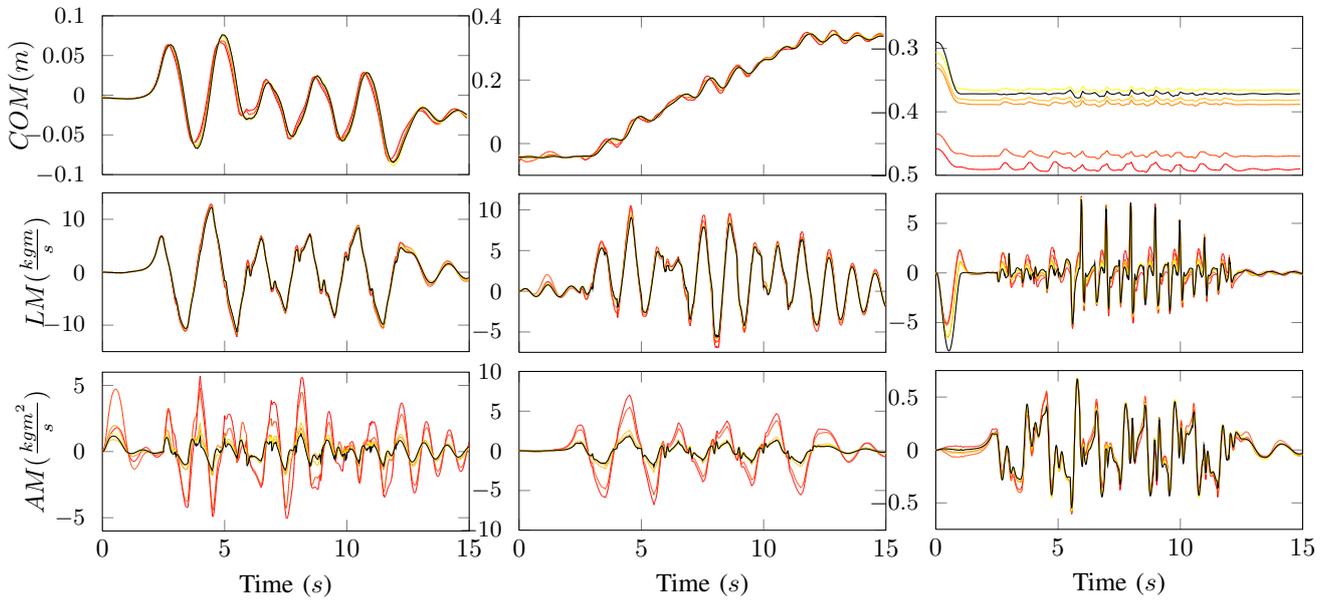

Fig. 6: Estimation of COM (top row), linear momentum (middle row) and angular momentum (bottom row) for different $n$ (representing different degrees of modeling error). Red denotes the most error while yellow denotes the least.

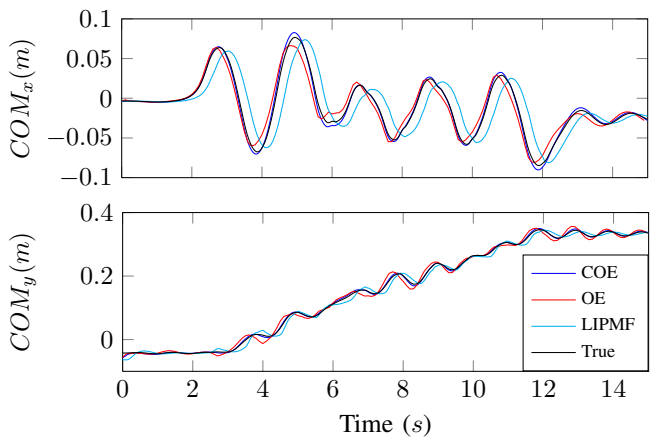

Fig. 7: COM estimation for $n = 5$ in $x$-direction (top) and $y$-direction (bottom).

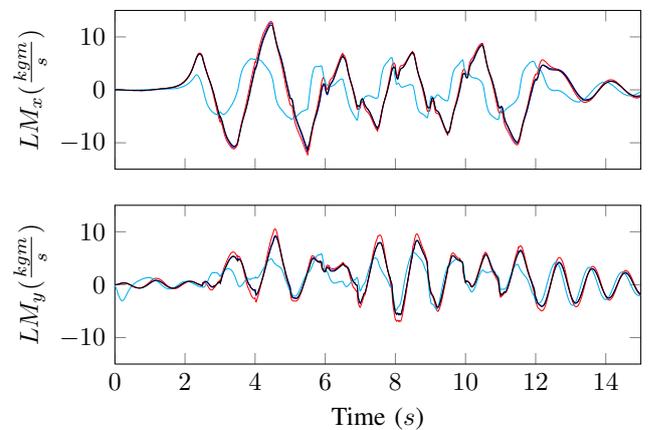

Fig. 8: Linear momentum estimation for $n = 5$ in $x$-direction (top) and $y$-direction (bottom).

TABLE II: Noise standard deviations for momentum-based estimators. N/A indicates that a parameter is not used.

| Name (Units) | ME | OE | COE | EWE |
|---|---|---|---|---|
| $q_{\Delta c}(m/\sqrt{Hz})$ | 1.0 | 1.0 | 1.0 | N/A |
| $q_{\Delta l}(kgm/s/\sqrt{Hz})$ | 1.0 | 1.0 | 1.0 | N/A |
| $q_{\Delta F}(N/\sqrt{Hz})$ | N/A | N/A | N/A | 1.0 |
| $q_{\Delta \tau}(Nm/\sqrt{Hz})$ | N/A | N/A | N/A | 0.1 |
| $r_c(m)$ | 0.0001 | 0.001 | 0.001 | 0.00001 |
| $r_l(kgm/s)$ | N/A | 1.0 | 1.0 | N/A |
| $r_k(kgm^2/s)$ | 0.1 | 1.0 | 10.0 | 0.01 |
| $r_{COP}(m)$ | N/A | N/A | 1.0 | N/A |

### A. Momentum Estimator

Both the LIPMF and ME accurately estimate the COM well as shown in Figure 1. This is expected since the COM is measured directly and $r_c$ was chosen small in both filters. However, linear momentum estimation is significantly better using the ME since the dynamics of this state are inaccurate in the LIPMF. The LIPMF relies on the COM measurement to correct for simplified model errors, introducing a delay.

The difference in performance between the two filters is more pronounced when they are updated at slower rates. Figure 2 shows estimation of $l_y$ by each estimator for update rates of $50Hz$, $125Hz$, $250Hz$, $500Hz$ and $1000Hz$ (lower frequencies are plotted in red, higher ones in yellow and ground truth in black). The degradation of performance with decreases in update rate is much more severe for the LIPMF.

The ME was motivated by the desire to avoid low-pass filtering computed momentum. Figure 3 shows the kinematics-based angular momentum filtered using cutoffs of 3% and 5% of the Nyquist frequency ($15Hz$ and $25Hz$ respectively for a $1kHz$ update rate) and compared to the estimated angular momentum. While the estimate lags by several milliseconds, the delays for cutoffs of $15Hz$ and $25Hz$ are as large as $20ms$ and $10ms$, respectively. For dynamic motions, low-pass filtering not only creates delays but can entirely change the structure of the signal.

### B. Offset Estimator

First, a constant offset of $\Delta c_x = 5cm$ was added directly to the COM measurement, as was done in [6] (assuming $\Delta l$ and $\Delta k$ remain zero). Both the LIPMF and the OE converge to the true COM, though the OE provides more accurate and responsive estimates as shown in the top plot of Figure 4.

Next, configuration-dependent offsets in the COM and momentum were created using a set of perturbed link parameters. The LIPMF manages to estimate the COM but with significant delay (up to $100ms$) as shown in the bottom plot of Figure 4. This delay is the result of using simplified dynamics and was unimproved by tuning LIPMF parameters.

The fact that the OE can track the COM and linear momentum offsets despite the unmodeled angular momentum offset suggests robustness to unmodeled errors. In order to analyze this we estimate offsets for five different sets of link mass parameters. Each was generated by adding to every mass-weighted link COM position $m_i x_{COM_i}$ an offset drawn from a zero-mean Gaussian having standard deviation $n(m_i x_{COM_i})$ with $n = 1, 2, 3, 4, 5$. Figure 5 shows the generated configuration-dependent COM offsets throughout the walking task, with red denoting the most perturbed parameters ($n = 5$). The COM offset is as large as $4cm$ in $x$ and nearly twice that in $y$ for $n = 5$. Each perturbed model also results in offsets in $c_z$ and momentum but these are not shown to save space. Figure 6 shows the performance of the OE for the different models; black denotes ground truth. Note that angular momentum estimation is poor yet COM and linear momentum estimation remain relatively accurate even for larger values of $n$, demonstrating OE robustness.

### C. COP-Based Offset Estimator

It is clear that performance of the OE is degraded for significant modeling errors due to the unmodeled angular momentum offset. By including the force-based COP measurement and tuning $r_{COP}$ against $r_k$, we achieve accurate COM and linear momentum estimation for $n = 5$ as shown in Figures 7 and 8. Also shown are the corresponding LIPMF estimates; the LIPMF estimates the COM fairly well but with significant delay while linear momentum estimation is poor due to the unmodeled offset which is significant for $n = 5$.

### D. External Wrench Estimator

We first estimate a constant external force equal to half the robot's weight as was done in [6]. This is applied at the left hip of the robot in the $y$-direction, creating associated torques about the COM in $x$ and $z$. The top row of Figure 9 shows that the EWE quickly converges to the true external force and COM torque. The LIPMF converges to the wrong value, likely because unlike in [6], the force is physically applied in simulation and there is realistic noise on the COP measurement. This could not be improved with tuning.

The second row of Figure 9 shows estimation of a $10N$ force applied in the same manner during the walking task. Despite the fact that the robot is subject to impulsive forces throughout, the EWE provides accurate estimates of both external force and the small, configuration-dependent COM torques. The LIPMF has difficulty estimating the value of the force due to impulses resulting from contact switching.

Finally, the bottom row of 9 shows estimation of a $50N$ disturbance applied at time $10s$ for $0.5s$ during the walking task. The EWE overshoots the force value but performs much better than the LIPMF overall, exhibiting a fast rise time. As shown in blue, EWE estimation during the transient disturbance is improved by increasing the external torque process noise. However, this comes at a price - estimates are seen to be much noisier.

## VI. DISCUSSION

We summarize the results of the previous section below.
- The ME and LIPMF both estimate the COM well but the ME performs better at estimating linear momentum.
- LIPMF performance degrades much more rapidly than ME performance for decreasing update rates.
- The ME filters the COM and momentum with much less delay than a typical low-pass filter.
- Both the OE and LIPMF estimate a constant COM offset well but the OE performs much better for configuration-dependent offsets since it estimates $\Delta l$.

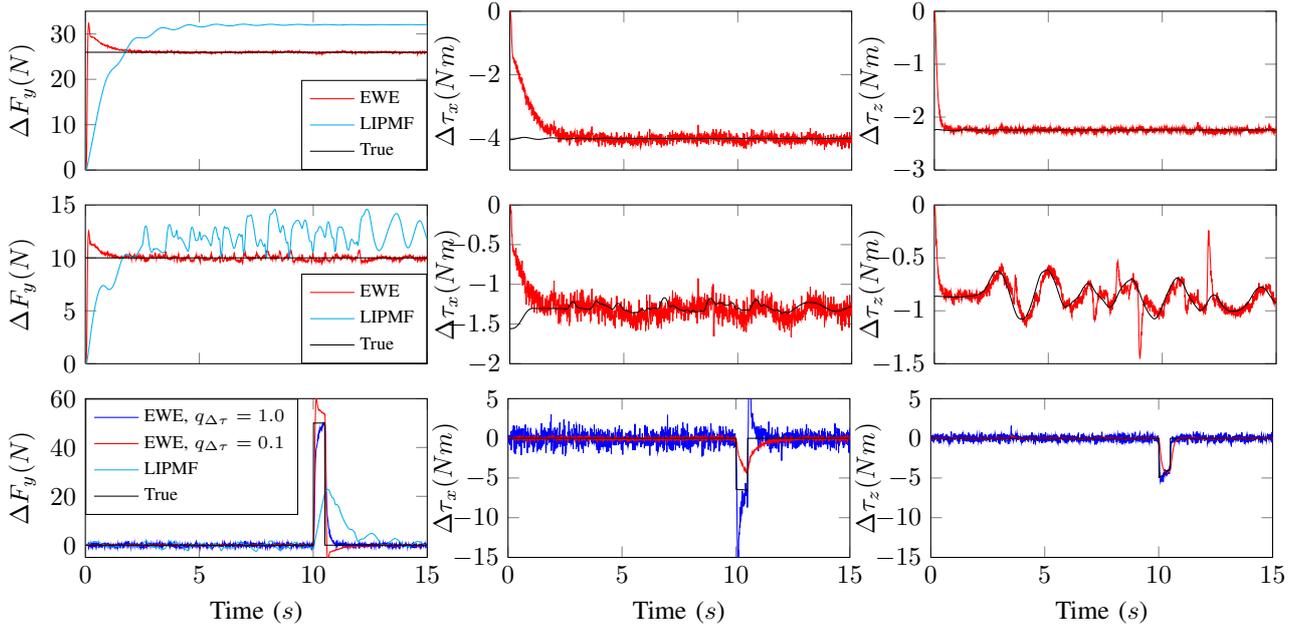

Fig. 9: External wrench estimation for a constant external force in global $y$-direction while stationary (top row), while walking (middle row) and for an impulsive force (bottom row) while walking (with process noise values of 0.1 and 1.0)

- The COE estimates offsets more accurately than the OE in cases of large modeling errors.
- The EWE accurately estimates constant, configuration-dependent and impulsive external wrenches even during the walking task.

Overall, the momentum-based filters perform better than the corresponding LIPM-based filters with only the disadvantage of being slightly more complicated to implement. The estimator to use depends on the application, though in practice we expect both offsets and external wrenches to exist on the real robot. Based on the analysis of Section IV-E we expect that a combined estimator may work well with proper tuning as long as the external COM torque is approximately zero. Alternatively, we may run both an offset estimator and an external wrench estimator in parallel and tune them accordingly. We plan to implement these approaches on the real robot and evaluate their performance in combination with the momentum control framework introduced in [4].

## VII. Conclusions

The momentum-based filter presented in this work estimates the COM and momentum of the system accurately and can be formulated into different filters depending on whether a COM offset or external wrench is affecting the robot. Other than the COE, none of the filters use the COP and are thus valid for any contact configuration (whereas the LIPMF assumes coplanar contacts). The momentum-based filter states were proven to be observable and were additionally shown to be robust to modeling errors as well as to slow update rates. These estimators are simple to implement and can be applied in a wide range of scenarios as demonstrated.